\documentclass[conference, a4paper]{IEEEtran}
\IEEEoverridecommandlockouts

\ifCLASSINFOpdf
  \usepackage[pdftex]{graphicx}
  \DeclareGraphicsExtensions{.pdf,.jpeg,.png}
\else
\fi

\usepackage{cite}
\usepackage{comment}
\usepackage{amsmath,amssymb,amsfonts}
\usepackage{algorithmic}
\usepackage{subcaption}
\usepackage{graphicx}
\usepackage{textcomp}
\usepackage{booktabs}
\usepackage{multirow}
\usepackage[left=1.29cm, right=1.29cm, top=1.9cm, bottom=4.29cm]{geometry}

\usepackage{optidef}
\usepackage{xcolor}
\usepackage{hyperref}
\usepackage{longtable}
\usepackage{gensymb}
\usepackage[]{algorithm2e}

\usepackage{glossaries}
\usepackage{soul}
\usepackage{flushend}

 \makeglossaries
\usepackage[font=small]{caption}
\usepackage{enumitem}
\usepackage{anyfontsize}
\usepackage{fancyhdr}
\makeatletter
\renewcommand{\fnum@figure}{Figure \thefigure}
\makeatother

\title{On the Generalization Properties of Deep Learning for Aircraft Fuel Flow Estimation Models
} 
\vspace{0.5cm}

\author{
\IEEEauthorblockN{Gabriel Jarry, Ramon Dalmau, Philippe Very}
\IEEEauthorblockA{
\small
EUROCONTROL \\
Brétigny-Sur-orge, France\\
\{gabriel.jarry, ramon.dalmau, philippe.very\}@eurocontrol.int}
\and
\IEEEauthorblockN{Junzi Sun}
\IEEEauthorblockA{
\small
Faculty of Aerospace Engineering \\
Delft University of Technology \\
Delft, the Netherlands\\
j.sun-1@tudelft.nl}
}

\IEEEaftertitletext{\vspace{-1\baselineskip}}

\begin{document}

\maketitle
\thispagestyle{fancy}

\begin{abstract}
Accurately estimating aircraft fuel flow is essential for evaluating new procedures, designing next-generation aircraft, and monitoring the environmental impact of current aviation practices. This paper investigates the generalization capabilities of deep learning models in predicting fuel consumption, focusing particularly on their performance for aircraft types absent from the training data. We propose a novel methodology that integrates neural network architectures with domain generalization techniques to enhance robustness and reliability across a wide range of aircraft. A comprehensive dataset containing 101 different aircraft types, separated into training and generalization sets, with each aircraft type set containing 1,000 flights. We employed the base of aircraft data (BADA) model for fuel flow estimates, introduced a pseudo-distance metric to assess aircraft type similarity, and explored various sampling strategies to optimize model performance in data-sparse regions. Our results reveal that for previously unseen aircraft types, the introduction of noise into aircraft and engine parameters improved model generalization. The model is able to generalize with acceptable mean absolute percentage error between 2\% and 10\% for aircraft close to existing aircraft, while performance is below 1\% error for known aircraft in the training set. This study highlights the potential of combining domain-specific insights with advanced machine learning techniques to develop scalable, accurate, and generalizable fuel flow estimation models.
\end{abstract}

\begin{IEEEkeywords}
Fuel Flow Estimation,
Aviation Sustainability,
Machine Learning,
Neural Network,
Domain generalization,
\end{IEEEkeywords}

\section{Introduction}
\label{sec:intro}

The aviation industry is a notable contributor to global CO2 emissions, responsible for approximately 2.5\% of the total -- a share expected to increase as air travel continues to expand~\cite{GOSSLING2020102194}. In addition to CO2, aviation generates non-CO2 effects, such as contrail formation, which further contribute to global warming~\cite{lee2021contribution}. As awareness of these environmental impacts grows, the industry has set ambitious goals to reduce its carbon footprint. International initiatives, such as the U.S. NextGen program and the Single European Sky Air Traffic Management Research (SESAR) Joint Undertaking, aim to significantly reduce aviation's environmental impact by 2035 through a variety of technological advancements and procedural innovations~\cite{undertaking2015european}.

Achieving these goals necessitates a multi-faceted approach, including advancements in engine technology, more aerodynamically efficient wing designs, and optimized air traffic management (ATM) procedures. For example, operational practices like continuous (cruise) climb operations (CCOs) and continuous descent operations (CDOs) can significantly cut fuel consumption and emissions~\cite{dalmau2015fuel, Clarke2004}. However, assessing the environmental benefits of such procedures depends on accurate and reliable fuel consumption models, which are also crucial for monitoring the environmental impact of operations.

Numerous models have been developed to estimate fuel consumption, each employing different methodologies and data sources. Many rely on mathematical approximations derived from basic physical equations or empirical data analysis~\cite{nuic2010bada, poll2021estimation}. The data to train these models ranges from widely accessible sources, like automatic dependent surveillance-broadcast (ADS-B)~\cite{sun2022openap}, to more detailed proprietary information from flight data recorders (FDR)~\cite{jarrytowards}. While these models are valuable, they come with limitations. Polynomial models, for instance, often oversimplify the complexities of fuel consumption, leading to potential inaccuracies in certain operational scenarios. For example, some models may provide excellent accuracy during the cruise phase, but not during the descent phase. 
Moreover, the simplicity of existing models typically necessitates the use of a separate model for each aircraft type. This fragmentation limits their applicability, especially in cases where data availability is sparse. If separate models are trained for each aircraft type, the data available for each may be insufficient to achieve high accuracy. Additionally, data for some aircraft types might be unavailable or only available for certain flight phases, such as cruise.

In response to these challenges, we propose a novel approach: the development of a generic and unique fuel consumption model that is conditioned on the specific characteristics of the aircraft, such as the wing span. This approach offers several key advantages over traditional models. First, by conditioning on aircraft characteristics, a single model can be applied across multiple aircraft types, eliminating the need to develop and maintain a separate model for each type. Second, the model can be trained on a larger and more diverse dataset, encompassing various aircraft types, thereby enhancing its robustness and accuracy. Third, and perhaps most importantly, this model has the potential to generalize across different scenarios, including those not represented in the training data. This capability, known as domain generalization in machine learning~\cite{zhou2022domain}, makes it a powerful tool for what-if analyses and for forecasting fuel consumption under new or unforeseen conditions, such as new aircraft designs or operational procedures.

This paper presents this model and, most critically, empirically evaluates its domain generalization properties. 

\section{State of the Art}
\label{sec:state}

This section presents the state of the art in two key areas. The first subsection reviews the latest advancements in fuel estimation techniques, while the second subsection explores the current approaches to domain generalization.

\subsection{Fuel Estimation}

Accurately estimating aircraft fuel consumption remains a significant challenge in ATM. To address this, researchers have developed various modeling techniques, ranging from traditional physical models to advanced neural networks and machine learning approaches. This subsection highlights the most relevant works across these different methodologies.

\subsubsection{Physical models} EUROCONTROL's Base of Aircraft Data (BADA) has been a widely recognized aircraft performance model that provides a foundational framework for simulating aircraft trajectories using total energy modeling. BADA encompassed comprehensive thrust, drag, and fuel consumption models for a broad spectrum of aircraft types~\cite{nuic2010user, nuic2010bada}. In pursuit of greater accessibility, OpenAP was designed specifically for the research community as an open-source model, promoting transparency and broad collaboration~\cite{sun2020openap}. Recently,~\cite{poll2021estimation, poll2021estimation2} proposed a method based on aerodynamic theory and empirical data to predict cruise fuel consumption and performance characteristics for turbofan aircraft.

\subsubsection{Machine learning models} Recent studies demonstrated the effectiveness of machine learning over conventional physical models. For instance, \cite{chati2017gaussian} used Gaussian process regression and operational data to enhance model accuracy. Building on this, the same authors predicted fuel flow rates using Classification And Regression Trees and least squares boosting, significantly improving emissions inventory accuracy~\cite{chati2016statistical}. Similarly, \cite{baumann2020modeling} leveraged full-flight sensor data with machine learning methods to outperform traditional fuel flow models. High accuracy in fuel flow prediction was further showcased by~\cite{baklacioglu2021predicting} through the use of advanced neural networks such as radial basis function networks. Evaluating neural network models against specific aircraft types,~\cite{trani_neural_2004} highlighted the potential for real-time simulation applications. Additionally, \cite{li2021study} employed long-short term memory (LSTM) neural networks to accurately predict performance-based contingency fuel.

In the context of exhaust emissions and combustion efficiency, \cite{kayaalp2021developing} achieved high accuracy with an LSTM model without extensive experimental testing. Likewise, \cite{metlek2023new} developed a model that surpassed previous methods in accurately predicting aircraft fuel consumption. The integration of Genetic Algorithm-optimized neural networks by \cite{baklacioglu2016modeling} introduced a novel approach to fuel consumption prediction across various flight phases. Moreover, \cite{uzun2021physics} proposed a hybrid strategy that incorporated physics-based loss during training to enhance model robustness against parameter changes, showing significant promise in making neural network models more resilient. In previous research, we benchmarked neural networks for estimating on-board aircraft parameters such as fuel flow and flap configuration during approach and landing ~\cite{jarry2020approach}, which could lead to improved ATM metrics ~\cite{jarry2021toward}. We also released an open-source generic aircraft fuel flow regressor for ADS-B data, trained on quick-access recorder data~\cite{jarrytowards}.

\subsection{Domain Generalization}

Accurate estimation of aircraft fuel flow is critical for several applications, including the evaluation of new procedures, the design of next-generation aircraft, and the assessment of the environmental impact of aviation. The challenge is to develop models that can generalize across different aircraft types and operating conditions, especially those not included in the training data. To address this challenge, the concept of domain generalization has emerged as a promising approach. This paper is among the first to specifically quantify the generalization capabilities of deep learning models in the context of aircraft fuel flow estimation.

Domain generalization aims to create models that perform well in unseen domains by using training data from multiple source domains \cite{zhou2022domain}. This is particularly relevant for aircraft fuel flow estimation, where models must generalize to different aircraft types and flight conditions. Domain alignment techniques are essential for this purpose, as they help minimize the differences between the source domains, allowing the model to learn domain-invariant representations \cite{muandet2013domain}, \cite{li2018domain}. 


Data augmentation plays a critical role in simulating domain shifts to improve model generalization. For aircraft fuel flow estimation, data augmentation can be used to create different training scenarios that mimic different aircraft behaviors, characteristics and environmental conditions. Techniques such as random augmentation networks \cite{xu2020robust} and feature-based augmentation \cite{zhou2021domain} can be particularly effective in expanding the diversity of training data, thereby improving the model's ability to generalize.

Finally, regularization strategies are critical to improving model robustness by reducing reliance on local features and encouraging the use of global structures. In the context of fuel flow estimation, this could involve regularizing the model to focus on general aerodynamic principles rather than specific data peculiarities of particular aircraft types. Techniques such as iteratively masking over-dominant features \cite{huang2020self} could be integrated with domain alignment and data augmentation to further improve performance.

While these general domain generalization techniques are valuable, context-specific approaches tailored to aircraft fuel flow estimation could offer even greater benefits. Inductive bias plays a critical role here by embedding domain-specific knowledge into the learning process. For example, physically-informed neural networks (PINNs) \cite{raissi2019physics, cuomo2022scientific} can incorporate physical laws and constraints, such as flight dynamics equations \cite{uzun2021physics}, directly into the model. This approach not only improves generalization, but also ensures that predictions are consistent with known physical principles, resulting in more reliable and robust fuel flow estimates over a wide range of aircraft and operating conditions.

The issue of robustness and generalisation of regression algorithms is also studied in classical statistics when sampling is required to build training data sets. Stratification techniques are used to optimally represent a population by dividing it into distinct subgroups called strata. Equal allocation, which ensures that each stratum has the same sample size, can increase the robustness of estimators by ensuring that under-represented strata are not overlooked \cite{singh1996stratified}, \cite{barnabas2014comparison}

Recently, a similar idea has been explored using an advanced reweighting scheme \cite{steininger2021density}. The authors adjust the influence of each data point using kernel density estimation (KDE), giving rare data points more influence on model training. This in turn allows for more robust estimators that are less prone to overfitting on redundant data in the training set.

\subsection{Contribution of this paper}

This paper aims to demonstrate and quantify the generalization properties of neural networks \cite{jarry2020approach, jarry2021toward} to accurately predict fuel flow even for unseen aircraft types by applying simple domain generalization techniques such as data augmentation and model regularization. 

First, we extended the aircraft features included in previous models \cite{jarrytowards}, by integrating more parameters from aircraft (wing span, reference masses or speeds etc.) and engines (by-pass ratio, rated thrust, reference fuel consumption etc.) characteristics.

Second, we aggregated a large representative dataset of 101,000 flights with 101 different aircraft types divided into two subsets: a primary data of 64 aircraft used for training, and a secondary dataset composed of 37 aircraft used for generalization assessment. We applied the BADA 4.2.1 model to obtain an estimate of fuel flow (this assumption and its implications are discussed further in section \ref{subsec:discussion}.). We defined a pseudo-distance between aircraft types, and applied a uniformization process to the data. 

Third, we conducted experiments to assess the generalization performance of the neural network on the unseen aircraft functions of their distance to the training dataset. To the best of our knowledge, this is the first paper to demonstrate and quantify the ability of fuel flow models to deal with unseen aircraft.

\section{Methodology}
\label{sec:process}

This section outlines the methodology used to develop a generic model and assess its generalization capabilities for predicting the fuel flow of  aircraft types not included in the training set. The process is presented in chronological order: Section~\ref{subsec:data} covers the data collection and preparation; Sections~\ref{subsec:distance} and~\ref{subsec:uniform} introduce the distance metric for evaluating aircraft type similarity and the uniformization process. Section~\ref{subsec:loss} presents the losses and training process, and Section~\ref{subsec:model} describes the neural network model architecture.

\subsection{Data collection and preparation}
\label{subsec:data}

To train and evaluate a machine learning model for fuel flow, a labeled dataset (i.e., with both inputs and outputs) of flight states (such as true airspeed, altitude, aircraft mass, etc.) and the corresponding fuel flow is required. Additionally, if a generic model applicable to various aircraft types is desired, the model must be conditioned on aircraft and engine-specific characteristics. These two types of data are described below.

\subsubsection{Flight Data}

ADS-B data were obtained from the OpenSky network~\cite{schafer2014bringing}, with a focus on flights conducted in 2022. The raw data were resampled at 4-second intervals using the \emph{traffic} library~\cite{olive2019traffic}, and all flight trajectories were enriched with weather data—specifically, wind and temperature—using the \emph{fastmeteo} library~\cite{sun2023fast}. To enhance the dataset, the derivatives of ground speed and true airspeed were calculated after applying an 8-second moving average smoothing.

It should be noted that the majority of the flights originate from Europe and the United States, regions with robust ground receiver coverage. For short- to medium-range flights, most selected trajectories include complete flight paths from takeoff to landing. However, for long-range flights, such as those operated by the Airbus A380, some segments may be missing in oceanic regions where receiver coverage is unavailable.

The features extracted to represent the flight state in the model include altitude (ft), vertical rate (ft/min), ground speed (kt), true airspeed (kt), ground acceleration (kt/s), air acceleration (kt/s), and air temperature (K). Additionally, the aircraft type was included to be matched with the corresponding aircraft and engine characteristics described below.

From this global dataset, 64 aircraft types were selected to form the primary dataset which is used for training the model, and 37 others to build the secondary dataset used to assess the generalization performance. Each aircraft type account for exactly 1,000 flights each. Since only ICAO type is available in the ADS-B data, we apply the full aircraft type BADA 4.2.1 model to randomly selected ADS-B trajectory of its corresponding ICAO type. To prevent data leakage, we ensured that ADS-B flights included in the secondary dataset were not present in the primary dataset (even for same ICAO type). The aircraft types included in each dataset are provided in Table~\ref{tab:primary_acft}.

\begin{table}[ht]
    \centering
    \scriptsize
    \caption{Aircraft lists for primary and secondary datasets}
    \begin{tabular}{p{8cm}}
        \toprule
        \textbf{Primary Dataset} \\
        \midrule
        A300B4-601, A330-321, B752WRR40, EMB-135LR, A300B4-622, A330-341, B753RR, EMB-145ER, A318-112, A350-941, B762ERPW56, EMB-145LR, A319-114, ATR42-500, B762GE50, EMB-145XR, A319-131, ATR72-200, B763ERGE61, EMB-170AR, A320-212, ATR72-500, B763PW60, EMB-170LR, A320-214, ATR72-600, B764ER, EMB-170STD, A320-231, B712HGW21, B772LR, EMB-175AR, A320-232, B73320, B772RR92, EMB-175LR, A320-271N, B73423, B773ERGE115B, EMB-175STD, A321-111, B737W24, B788RR67, EMB-190AR, A321-131, B738W26, B788RR70, EMB-190LR, A330-203, B739ERW26, B789GE75, EMB-190STD, A330-223, B744ERGE, B789RR64, EMB-195AR, A330-243, B744GE, B789RR74, EMB-195LR, A330-301, B748F, EMB-135ER, EMB-195STD \\
        \midrule
        \textbf{Secondary Dataset} \\
        \midrule
        A300B4-203, A340-642, B742RR, F100-650, A300B4-608ST, A380-841, B743PW, F70-620, A310-204, A380-861, B773RR92, MD808120, A310-222, ATR42-300, B788GE67, MD808221, A310-308, ATR42-320, B788GE70, MD808321, A310-322, ATR42-400, B788RR53, MD808720, A310-324, ATR72-210, B788RR64, MD808821, A340-213, B73215, EMB-135BJ-L600, A340-313, B73518, EMB-135BJ-L650, A340-541, B73622, F100-620 \\
        \bottomrule
    \end{tabular}
    \label{tab:primary_acft}
\end{table}

For each observation in both the primary and secondary datasets, we calculated the estimated fuel flow for six different takeoff masses, ranging from 70\% to 95\% of the maximum takeoff weight (MTOW), in 5\% increments, using the BADA 4 model (via the \emph{pyBADA} library)~\cite{dalmau2018pybada}.

Finally, both datasets (primary and secondary) were randomly split into training, validation, and testing subsets using an (80\%, 10\%, 10\%) distribution on flight basis, ensuring the independence of each subset. The train and validation subsets of primary dataset are used to train and select the model, the test subset is use to assess the performance on known aircraft. The train and validation subsets of secondary dataset are not used at all in the following experiments (they are prepared to build a full model), only the test subset of secondary dataset is used to assess the generalization performance on unseen aircraft.

\subsubsection{Aircraft and engine characteristics}



To create a model that generalizes across a wide range of aircraft types, aircraft and engine characteristics were matched to each observation according to the corresponding aircraft type. Table~\ref{tb:aicraft_characteristics} provides details on the features representing the aircraft and engine characteristics included in each dataset observation. 

\begin{table}[h!]
\centering
\caption{Aircraft and engine features. Aircraft characteristics were obtained from open data sources, while engine features were sourced from ICAO, FOCA, and FOA emissions databases.}
\label{tb:aicraft_characteristics}
\begin{tabular}{lll}
\toprule
\textbf{Type} & \textbf{Feature} & \textbf{Unit} \\
\midrule
\multirow{8}{*}{ADS-B \& weather} & Altitude & ft \\
& Vertical Rate & ft/min \\
& Ground Speed & kt \\
& True Air Speed & kt \\
& Ground Acceleration & kt/s \\
& Air Acceleration & kt/s \\
& Temperature & K \\
\midrule
\multirow{8}{*}{Aircraft} & Wing Area & $m^2$ \\
& Mass & kg \\
& Span & m \\
& Length & m \\
& Maximum Take-Off Weight (MTOW) & kg \\
& Zero Empty Weight (OEW) & kg \\
& Maximum Mach Operating (MMO) & mach \\
& Velocity Maximum Operating (VMO) & kt \\
& Altitude Maximum Operating (HMO) & ft \\
\midrule
\multirow{10}{*}{Engine} & Engine Type & - \\
& Number of Engines & - \\
& Rated Power & hp \\
& Rated Thrust & kN \\
& Bypass Ratio & - \\
& Pressure Ratio & - \\
& Take-off Fuel Flow & kg/s \\
& Climb Fuel Flow & kg/s \\
& Approach Fuel Flow & kg/s \\
& Idle Fuel Flow & kg/s \\
\bottomrule
\end{tabular}
\end{table}

It is important to note that not all characteristics are available for every aircraft type. When specific data is missing, it is imputed using linear regression iterative imputation from \emph{scikit-learn}~\cite{pedregosa2011scikit}.

\subsection{Distance metric for aircraft type similarity}
\label{subsec:distance}

In order to create a similarity measure between aircraft types, we applied a uniform \texttt{QuantileTransformer} from \emph{scikit-learn} on both aircraft and engine characteristics to deal with features with different ranges of values. This method transforms the features to follow a uniform  distribution, also reducing the impact of  outliers and therefore acting as a robust preprocessing scheme. The similarity measurement is defined as the $\ell_2$ norm in the uniform space. Examples of distance are displayed in Table~\ref{tab:distances}

\begin{table*}[h!]
\centering
\caption{Examples of computed distances between a subset of aircraft types.}
\begin{tabular}{lccccccccc}
\toprule
\textbf{Aircraft Type} & \textbf{A318-112} & \textbf{A319-114} & \textbf{A320-214} & \textbf{A330-243} & \textbf{A340-213} & \textbf{A350-941} & \textbf{A380-841} & \textbf{ATR42-400} & \textbf{ATR72-500} \\
\midrule
\textbf{A318-112} & 0.0 &  &  &  &  &  &  &  &   \\
\textbf{A319-114} & 0.11 & 0.0 &  &  &  &  &  &  &   \\
\textbf{A320-214} & 0.45 & 0.39 & 0.0 &  &  &  &  &  &   \\
\textbf{A330-243} & 1.82 & 1.78 & 1.53 & 0.0 &  &  &  &  &   \\
\textbf{A340-213} & 1.68 & 1.64 & 1.44 & 0.88 & 0.0 &  &  &  &   \\
\textbf{A350-941} & 1.98 & 1.93 & 1.66 & 0.73 & 0.91 & 0.0 &  &  &   \\
\textbf{A380-841} & 2.29 & 2.24 & 2.00 & 0.97 & 0.79 & 0.66 & 0.0 &  &   \\
\textbf{ATR42-400} & 1.69 & 1.72 & 1.95 & 3.13 & 2.93 & 3.32 & 3.60 & 0.0 &   \\
\textbf{ATR72-500} & 1.58 & 1.60 & 1.82 & 2.97 & 2.77 & 3.16 & 3.44 & 0.21 & 0.0 \\
\bottomrule
\end{tabular}
\label{tab:distances}
\end{table*}

\begin{figure}[ht!]
    \centering
    \includegraphics[width=0.85\linewidth]{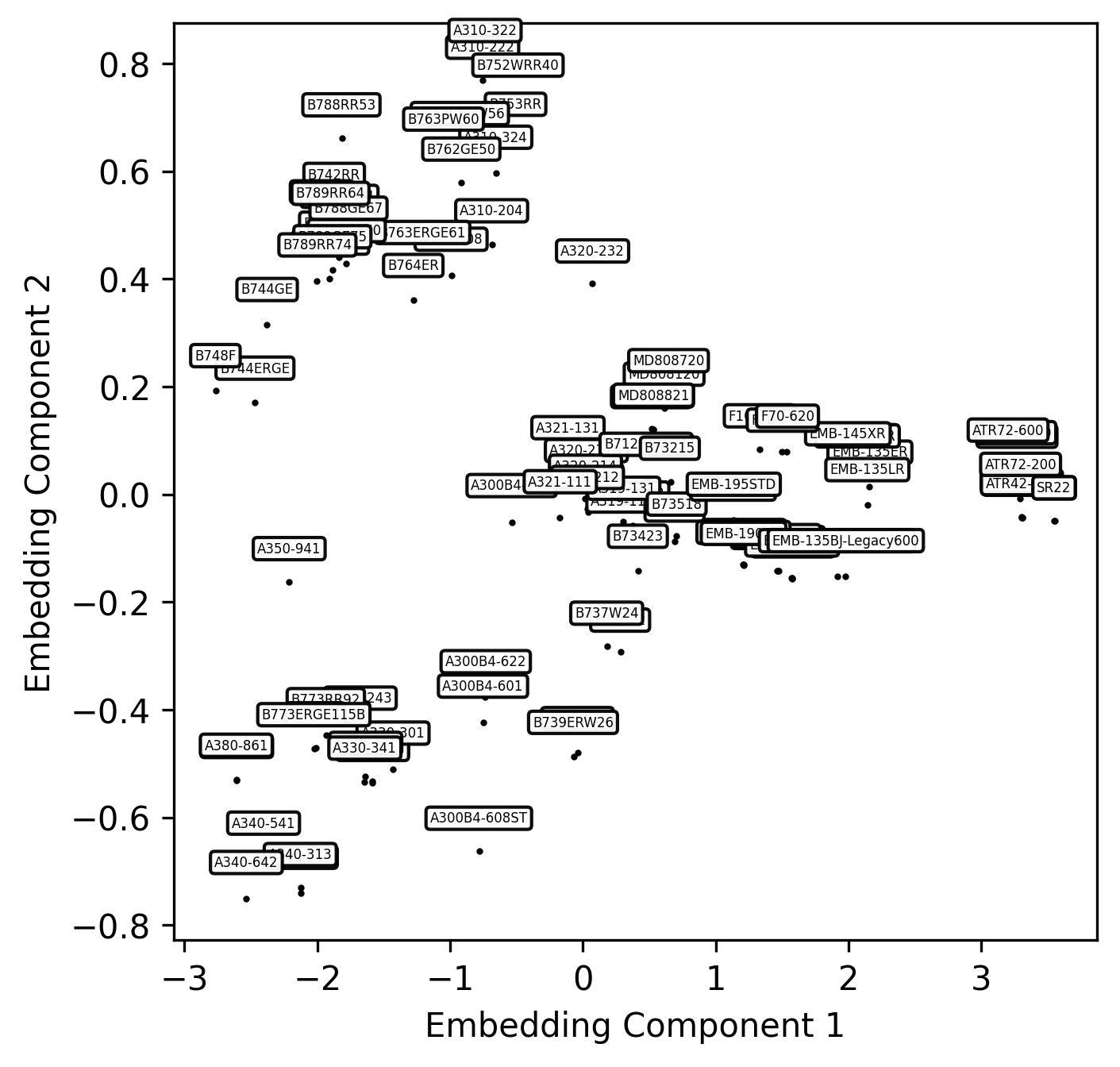}
    \caption{Illustration of the BADA 4.2.1 aircraft and engines embedded with the \texttt{Isomap} process.}
    \label{fig:embbeding}
\end{figure}

Figure~\ref{fig:embbeding} illustrates the aircraft embedding in two dimensions using an \texttt{Isomap} with 10 neighbors and 2 components, applied for illustrative purposes. The \texttt{Isomap} method~\cite{tenenbaum2000global} reduces dimensionality by preserving geodesic distances between data points, effectively capturing the intrinsic geometry of the data manifold. 


\subsection{Uniformization process}\label{subsec:uniform}

To analyze the model's generalization properties, we compare two types of sampling on both the training and validation sets, while keeping the test set unchanged.


First, we use random sampling with replacement: 1000 observations per flight trajectory for the training set and 500 for the validation set, drawn from the entire dataset.

\begin{figure}[ht!]
    \centering
    \includegraphics[width=1.\linewidth]{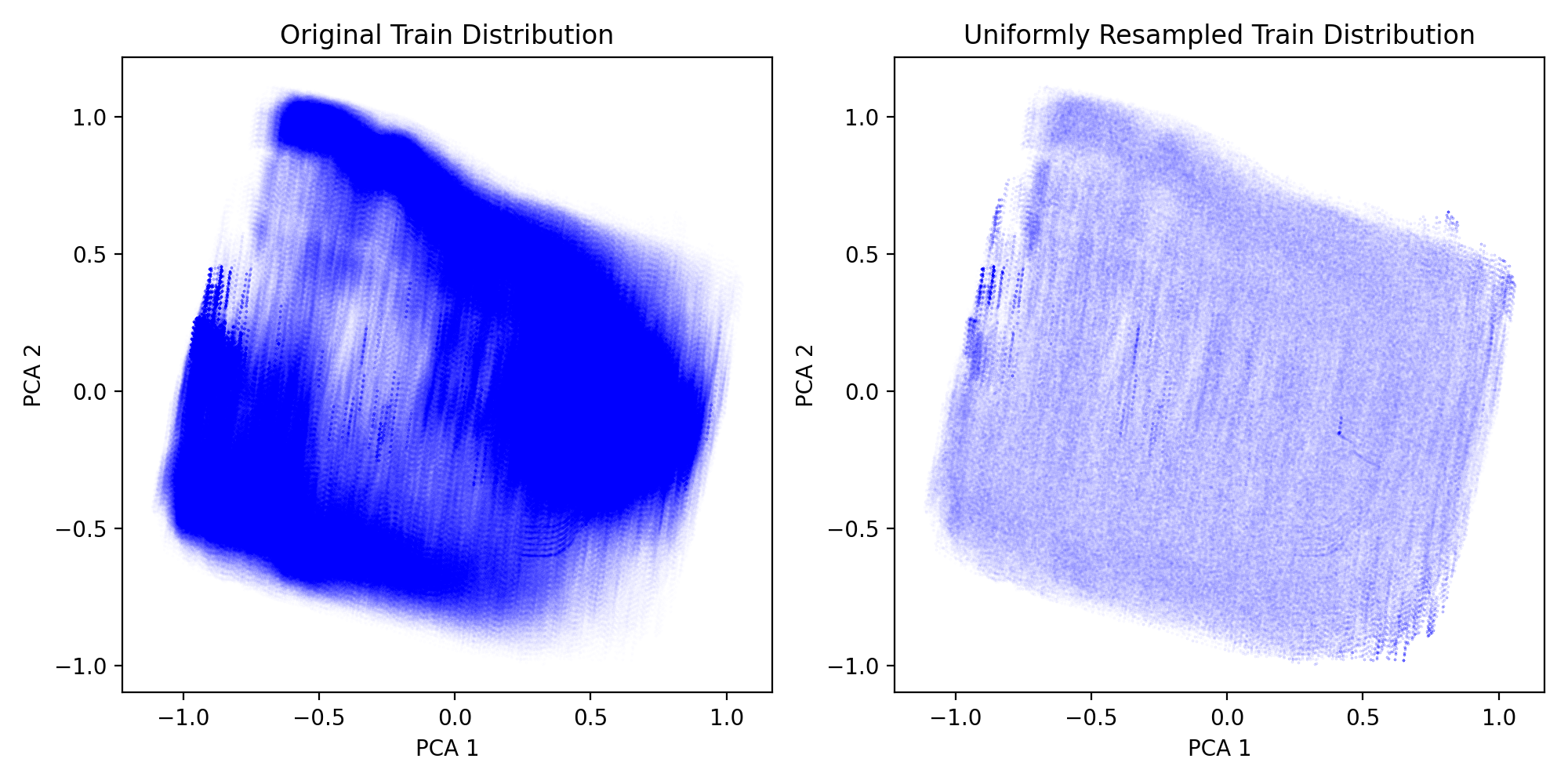}
    \caption{Example of sample train  set distribution in PCA latent space. On the left is the original train distribution, on the right is the distribution after the uniformization process.}
    \label{fig:uniformization}
\end{figure}

Second, we propose a uniform sampling method applied to the ADS-B parameters for each aircraft type. This process begins by applying again a uniform \texttt{QuantileTransformer} to the data, followed by dimensionality reduction using Principal Component Analysis (PCA) with 2 components, which projects each observations into a two-dimensional space. Kernel Density Estimation (KDE) is then used to estimate the density of observations in this latent space, with the weights for sampling calculated as the inverse of their estimated density, thus favoring observations in less dense regions. Dimension reduction is critical here to avoid the curse of dimensionality for density estimation. The result of this process is shown in Fig.~\ref{fig:uniformization}.

Finally, weighted sampling is performed based on these weights to select a uniform subset of the data according to the specified budget: 1000 observations per trajectory for training and 500 observations per trajectory for validation.

\subsection{Loss and training process}
\label{subsec:loss}

We trained all models on GPUs (NVIDIA RTX A4500 or RTX A6000 Ada Gen) using Tensorflow for 1K epochs with a checkpoint on the validation set and evaluated the performance of the models on the test set. To ensure reproducibility, we set the same seed in all experiments (ramdom and np.random seed: 42, tf deterministic ops: 1 and python hash seed: 0).

Different metrics are used to train or evaluate the performance of the model: the Mean Square Error (MSE), the Mean Error (ME), the Mean Absolute Error (MAE) and the Mean Absolute Percentage Error (MAPE). 

Additionally, we defined a combined loss between MAE and MAPE. Let $\mathcal{D}$ be a set of input-output pairs $(x,y)$ and $h$ be a model to evaluate, the metric is calculated as follows:
\begin{equation}
    \mathbf{\beta-MAPE}(h,\mathcal{D}) = \mathbf{MAE}(h,\mathcal{D}) + \beta\,\mathbf{MAPE}(h,\mathcal{D})
\end{equation}

\subsection{Model Architecture}\label{subsec:model}

\begin{figure}[ht!]
    \centering
    \includegraphics[width=0.8\linewidth]{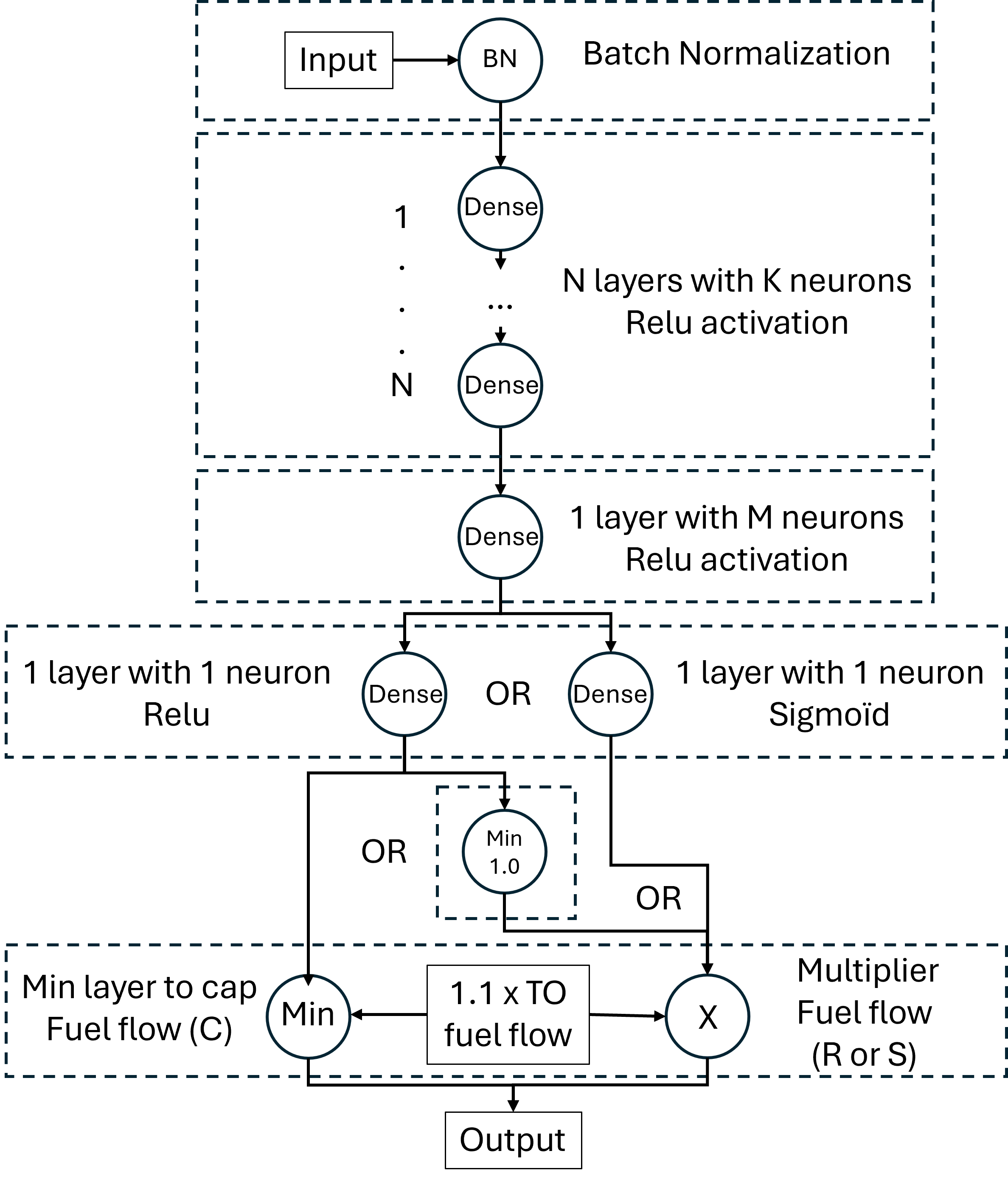}
    \caption{N-K-M dense architecture. The final layer has three variations. Variation (C) consists of a Relu activation and a min layer with 1.1 times fuel flow take-off. Variation (R) consists of a Relu activation, a minimum layer with 1.0 and finally a multiplication layer with 1.1 fuel flow take-off. Finally, variation (S) uses a sigmoid layer and a multiplication layer with 1.1 fuel flow take-off.}
    \label{fig:architecture}
\end{figure}

The proposed neural network architecture is an \(n\)-\(k\)-\(m\) dense neural network designed for robust and efficient performance. The model begins with a batch normalization layer to standardize the inputs, followed by a block of \(N\) fully connected layers, each comprising \(K\) neurons. These layers employ ReLU activation functions and are regularized with an $\ell_2$ kernel regularization coefficient of \(1 \times 10^{-4}\). After this block, a single dense layer with \(M\) neurons is added, also using ReLU activation and $\ell_2$ regularization. The final layer has three variations. Variation (C) consists of a Relu activation and a min layer with 1.1 times fuel flow take-off. Variation (R) consists of a Relu activation, a minimum layer with 1.0 and finally a multiplication layer with 1.1 fuel flow take-off. Finally, variation (S) uses a sigmoid layer and a multiplication layer with 1.1 fuel flow take-off. The architecture of the model is shown in Fig.~\ref{fig:architecture}.

\section{Results}
\label{sec:results}

In order to build the reference model we first performed a hyperparameter search to find a good configuration. It is not a complete grid search, each training last between 2h to 6h and the objective is not to find the best model just a good reference architecture. The selected architure is a 7-250-4 model trained with a 20-MAPE loss. All the final layer variations are tested in this section.

\subsection{Sampling process}

The analysis of the performance metrics on the test dataset in Table \ref{tab:sampling} shows that the choice of validation set sampling, whether uniform or random, does not impact the final model selection, as evidenced by identical MAPE, MAE, and ME values for both methods implied same model is selected with the check-pointing. However, the sampling method of the training set seems to influence performance, with uniform sampling yielding slightly better results. Models trained with uniform sampling achieve a lower MAPE of 0.54\% than with the random sampling with 0.67\% MAPE.

\begin{table}[h!]
\centering
\caption{Performance of the model on the test dataset for \textbf{different train and validation sampling}. The metrics are averaged over all the aircraft types, and the standard deviation is displayed under parenthesis.}
\label{tab:sampling}
\begin{tabular}{ccccc}
 \toprule
{Train} & {Validation} & {MAPE (\%)} & {MAE (kg/h)} & {ME (kg/h)} \\
 \midrule
\textbf{Uniform} & \textbf{Uniform} & 0.54 (0.10) & 9.08 (7.07) & 2.55 (3.51)\\
Uniform & Random & 0.54 (0.10) & 9.08 (7.07) & 2.55 (3.51)\\
Random & Uniform & 0.67 (0.13) & 9.92 (6.52) & 0.15 (3.37)\\
Random & Random & 0.67 (0.13) & 9.92 (6.52) & 0.15 (3.37)\\
 \bottomrule
\end{tabular}
\end{table}

\begin{figure}[ht!]
    \centering
    \includegraphics[width=0.9\linewidth]{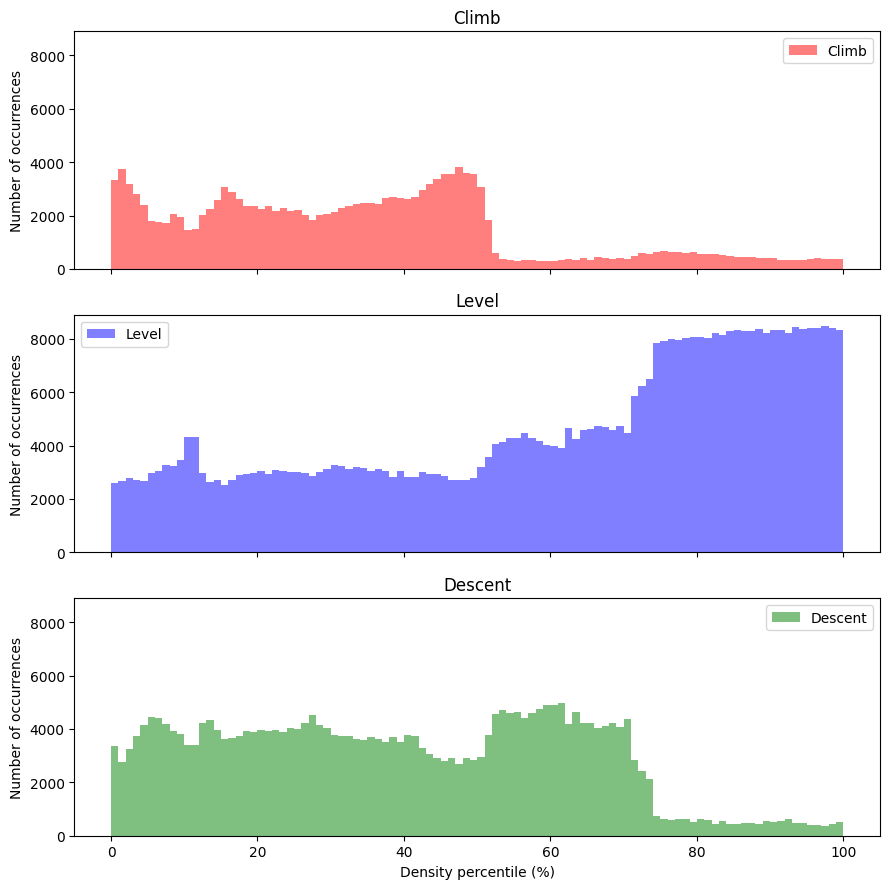}
    \caption{Histogram densities per phases}
    \label{fig:hist_densities}
\end{figure}

The uniform sampling strategy defined above aims to reduce the impact of dense (and therefore redundant) observations, which occur mainly during the level phase, as shown in Figure \ref{fig:hist_densities}. The uniformization process would likely act as a re-weighting of the flight attitude and can be beneficial if the goal is to improve performance during ascent and descent, which are generally the more difficult phases to estimate fuel flow \cite{jarrytowards}.

\subsection{Generalization performance}

The generalization property of the model is evaluated using the same metrics on the secondary dataset (37 unseen aircraft types). To ensure stability in training and testing, we will only change the architecture of the model here and see how the generalization property evolves. The task we are attempting is particularly challenging due to the design of our dataset, which excludes entire families of aircraft from the training set (A310, A340, A380, and the F and MD series). We aim to generalize the model's performance to these aircraft families unseen during training. In a more typical scenario, one would expect the model to have been exposed to at least one aircraft from the same family, making it a matter of predicting the performance of a new version or variant of that aircraft. This lack of representations in the training set significantly increases the complexity of the task, as the model must extrapolate from characteristics and behaviors it has never encountered before. Our study aims to demonstrate and quantify the model's ability to overcome this challenge and provide accurate predictions even under these adverse conditions.

\subsubsection{Last layer variation}

\begin{figure}[ht!]
    \centering
    \includegraphics[width=0.9\linewidth]{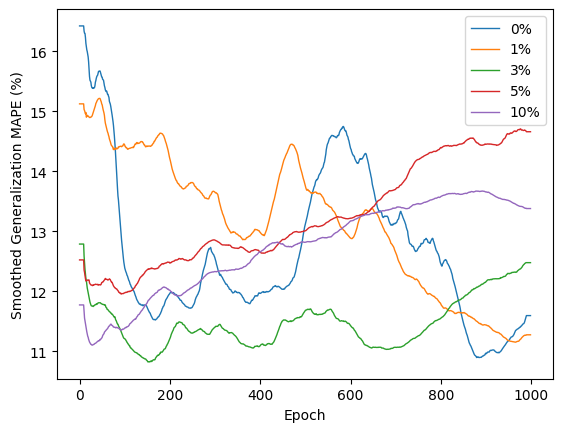}
    \caption{Generalization MAPE Performance (\%) curve during training process for different noise levels}
    \label{fig:gen_loss}
\end{figure}

Regarding the architectures in table \ref{tab:variation_gen}, we observe that the Relu (R) architecture presents the best results with 11\% MAPE, but if we check the evolution of the generalization performance during the learning process, we observe that it is really chaotic (blue curve, as shown in Figure \ref{fig:gen_loss}). In particular, we see that during training, the MAPE on the secondary data set increases and decreases, indicating a possible overfitting with respect to the aircraft and engine characteristics, while the validation curves (on the primary data set) show no sign of overfitting. Since the checkpoint is applied on the validation set we do not have any insurance that the learning stops during bad generalization performance.

\begin{table}[h!]
\centering
\caption{Performance of the model for \textbf{different last layer variation} on the secondary dataset. The metrics are averaged over all the aircraft types. The standard deviation is shown in parentheses.}

\begin{tabular}{cccc}
 \toprule
{Last Layer} & {MAPE (\%)} & {MAE (kg/h)} & {ME (kg/h)} \\
 \midrule
C & 20.39 (17.26) & 283.27 (224.26) & -9.52 (239.20)\\
\textbf{R} & 13.46 (10.07) & 220.09 (185.49) & 88.20 (208.14)\\
S & 18.69 (20.76) & 249.99 (246.19) & -95.44 (265.22)\\
\bottomrule
\end{tabular}
\label{tab:variation_gen}
\end{table}

\subsubsection{Adding noise}

To try to mitigate this behavior, we applied some noise regularization to the input aircraft and engine features. For each feature, we sampled a Gaussian distribution with a mean \(\mu = 0\) and a standard deviation \(\sigma = 0.33\).  Each noisy feature \(\hat{X}_p\) is computed as

\[
\hat{X}_p = X \cdot \left(1 + p \cdot \mathcal{N}(0, 0.33^2)\right)
\]

where \(X\) is the original feature and \(p\) is the noise percentage parameter. As observed in table \ref{tab:noise}, introducing noise improves model performance, especially at noise levels of 1\% and 3\%, where both MAPE and MAE show significant reductions. We also observe more stability in the MAPE curve during training and in particular with 1\% noise a proper decrease of the MAPE curve without overffiting periods.

\begin{table}[h!]
\centering
\caption{Performance of the model \textbf{with or without input noise} on the secondary dataset. The metrics are averaged over all the aircraft types, and the standard deviation is displayed under parenthesis.}
\label{tab:noise}
\begin{tabular}{ccccc}
 \toprule
{Noise p}  & {MAPE (\%)} & {MAE (kg/h)} & {ME (kg/h)} \\
 \midrule
0\% & 13.46 (10.07) & 220.09 (185.49) & 88.20 (208.14)\\
\textbf{1\%} & 10.34 (5.57) & 181.63 (162.72) & 40.40 (175.82)\\
3\% & 10.96 (6.13) & 183.96 (166.75) & 59.65 (169.79)\\
5\% & 13.00 (9.66) & 212.33 (179.87) & 71.07 (186.44)\\
10\% & 12.15 (10.88) & 207.86 (192.69) & 45.46 (204.17)\\

 \bottomrule
\end{tabular}
\end{table}

\begin{figure}[ht!]
    \centering
    \includegraphics[width=\linewidth,trim={0 0 0 1.2cm},clip]{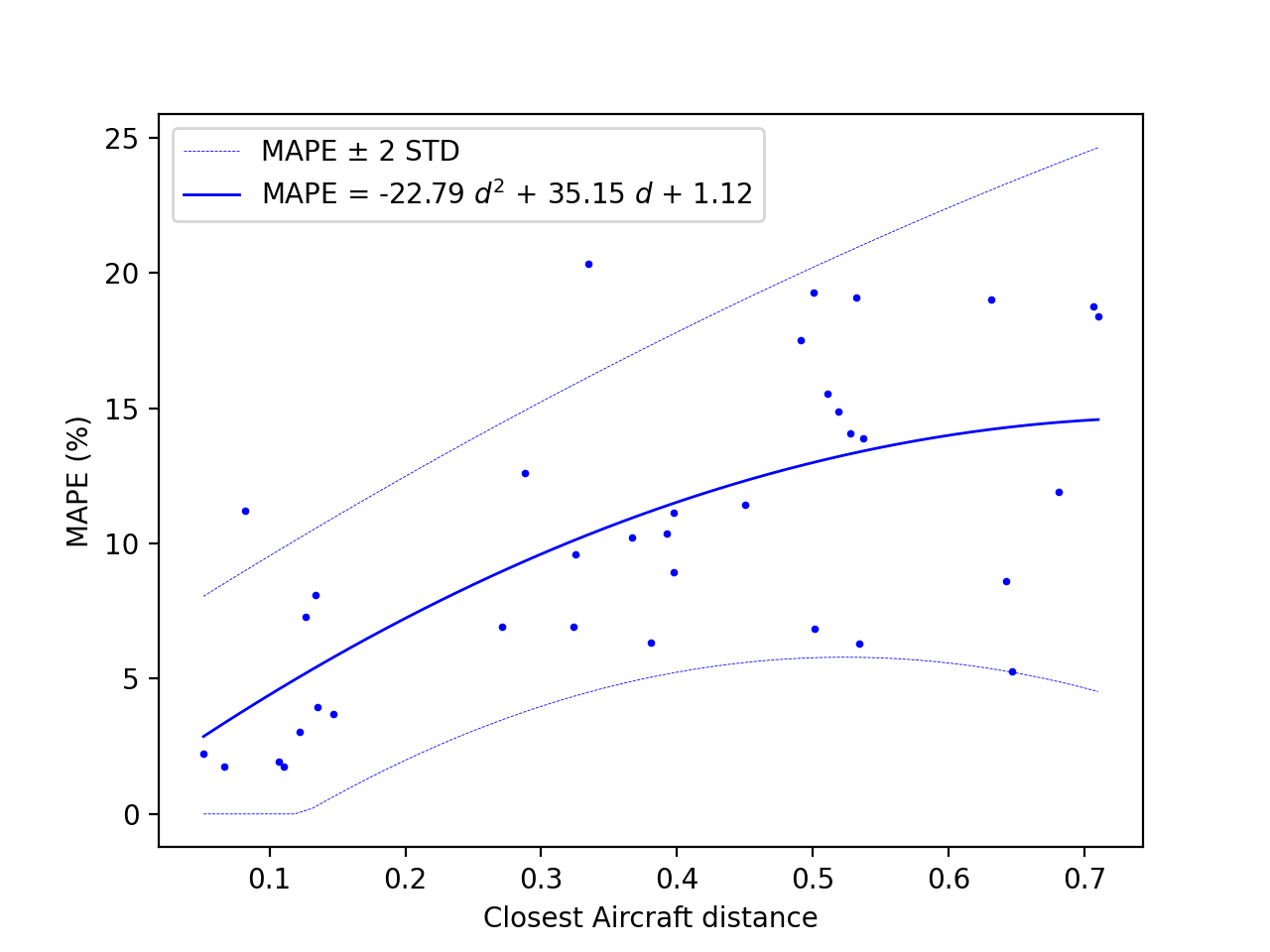}
    \caption{MAPE Performance (\%) of the best model (with 1\% noise) on secondary dataset function of the closest aircraft in the primary dataset. Each point is one of the 37 aircraft in the secondary dataset}
    \label{fig:gen_per_distance}
\end{figure}

Overall, we observe a linear/quadratic trend of the performance MAPE function of the distance to the closest aircraft in the primary data set in Figure \ref{fig:gen_per_distance}.  The error ranges from less than 5\% to 20\%. This is a really good property that implies the potential use of such a model to generalize to unseen aircraft with a quantified error measure.

\section{Discussion}
\label{subsec:discussion}

One limitation of our current approach is the discrete sampling of mass values. This method could be improved by implementing a random sampling technique for mass values, which would provide a more continuous representation. In addition, a more appropriate procedure would be to use historical mass distributions specific to each city pair. This improvement would allow for more accurate and realistic modeling of aircraft fuel flow that more closely reflects actual operational conditions. In addition, the current process is limited to 1,000 trajectories. Improving this by maximizing the variability within those flights could result in a more robust model that captures a wider range of flight conditions.

The uniform sampling method used here was able to improve the test performance. This can be explained by the nature of the input data distribution (redundancy of level phase data points with a large number of short and medium range aircraft). The performance could be probably increased by using active sampling strategies \cite{mindermann2022prioritized}, where the performance of the regression algorithm is fed back into the sampling algorithm to further improve its accuracy

An interesting finding relates to the overfitting observed on the aircraft and engine parameters. The model does not appear to overfit on the ADS-B parameters, where it is exposed to millions of data points, indicating robust generalization capabilities in this context. However, there is a noticeable tendency to overfit on the aircraft and engine parameters, which are limited to only 64 possible values. This discrepancy suggests that the model struggles with the limited variability of these features, leading to overfitting. Interestingly, introducing noise into the aircraft and engine parameters effectively alleviated this problem. By adding noise, we increase variability and prevent the model from learning spurious patterns specific to the training set, thereby improving its generalization performance to unseen aircraft types.

Further work could be done on the model architecture to introduce more inductive bias and improve generalization. For example, training an embedding through adversarial learning could help the model learn more robust representations. Additionally, learning the fuel flow distribution functions specific to each aircraft could provide more context-aware predictions. These approaches could help to better capture the underlying patterns in the data, thus improving the overall performance and robustness of fuel flow estimation models.

A major limitation of our current approach is the reliance on the BADA physical model as a proxy for estimating fuel flow. In reality, our goal is to develop and release models trained on Quick Access Recorder (QAR) data directly from the aircraft. The use of BADA provides a baseline, but it does not capture the full range of operational variability and subtleties present in actual flight data. The fuel data estimated using BADA appears to be significantly less noisy, which contributes to the exceptionally good performance (with errors less than 1\%) observed in our test results. The transition to models trained on QAR data will provide more accurate and reliable fuel flow estimates that better reflect real-world conditions.

\section{Conclusions}
\label{sec:conclusion}

In this paper, we explored the generalization properties of neural networks for aircraft fuel flow estimation, focusing on their ability to predict fuel flow for aircraft types not present in the training data. Our experiments produced several important results. First, we found that neural networks, when trained with the right architecture and hyper-parameters, can achieve highly accurate fuel flow predictions for observed aircraft types, with mean absolute percentage errors (MAPE) below 1\%.

Second, the effect of sampling methods was significant. Uniform sampling of the training data improved model performance during the ascent and descent phases, which are more complex and typically underrepresented in the raw data.

Third, generalization to unseen aircraft types was inherently more difficult. However, the introduction of noise in the aircraft and engine parameters significantly improved the robustness of the model, with mean absolute percentage errors around 10\%. This regularization technique prevented the model from overfitting to the limited set of aircraft and engine parameters, thereby improving its ability to generalize.

In our error analysis, we observed that the model's performance on the secondary dataset, consisting of unseen aircraft types, showed an increasing relationship between the prediction error and the distance to the closest aircraft type in the training dataset. This quantifiable error measure is critical for practical applications, as it allows stakeholders to understand and anticipate the model's limitations.

Looking ahead, our study highlights the potential benefits of incorporating domain-specific knowledge, such as physical constraints and historical mass distributions, into the training process. Future work should focus on refining these techniques and transitioning from proxy models such as BADA to direct use of Quick Access Recorder (QAR) data, which would provide more accurate and comprehensive datasets.

In addition, a complete BADA 4.2.1 (all 103 aircraft types) surrogate model has been built. Discussions are underway to release the model under a BADA license with an implementation in the DeepEnv library \cite{jarry2024deepenv}.

\bibliographystyle{ieeetr}

\bibliography{biblio.bib}

\end{document}